%% file: DeepCert.tex
\documentclass[runningheads]{llncs}
\usepackage{graphicx}
\usepackage{soul}
\usepackage{rotating}
\usepackage{graphicx}
\usepackage{amsmath}
\usepackage{amssymb}
\usepackage{blindtext}
\usepackage{makecell}
\usepackage{colortbl}
\usepackage{subcaption}
\usepackage{amsmath}
\usepackage[utf8]{inputenc}
\usepackage{csquotes}
\usepackage{cellspace}
\usepackage{array, tabularx, caption, boldline}
\usepackage{enumitem}
\usepackage{xcolor}
\usepackage{multirow}
\usepackage{booktabs}
\usepackage{bbding}

\usepackage{algorithm} 
\usepackage{algpseudocode} 

\usepackage{multirow,tabularx}

\usepackage{subcaption}
\usepackage{caption}
\usepackage{varwidth}

\usepackage{bbm}

\usepackage{hyperref}

\usepackage{todonotes}

\newcommand{\acronym}{DeepCert}

\newcommand{\vect}[1]{\boldsymbol{#1}}

\begin{document}
\title{\acronym: Verification of Contextually Relevant Robustness for Neural Network Image Classifiers}
\titlerunning{Contextually Relevant Robustness}
%
\author{Colin Paterson\inst{1} \Envelope \and Haoze Wu\inst{2} \and John Grese\inst{3} \and
 Radu Calinescu\inst{1}  \and \\ Corina S. P\u{a}s\u{a}reanu\inst{3} \and Clark Barrett\inst{2}}
\authorrunning{Paterson et al.}
%
\institute{
University of York, York, United Kingdom\\
\Envelope \email{colin.paterson@york.ac.uk} \and
Stanford University, Stanford, USA \and
Carnegie Mellon University, Silicon Valley, USA
}

\maketitle              

\input{sections/abstract}
\input{sections/introduction.tex}

\input{sections/method.tex}

\input{sections/implementation.tex}
\input{sections/experimental}

\input{sections/cifar}
\input{sections/related.tex}
\input{sections/conclusions.tex}

\vspace{0.5cm}
{\bfseries\noindent Acknowledgements.} 
This research has received funding from the Assuring Autonomy International Programme project `Assurance of Deep-Learning AI Techniques' and the UKRI project EP/V026747/1 `Trustworthy Autonomous Systems Node in Resilience'.

	\vspace{-1mm}

\bibliographystyle{splncs04}
\bibliography{DeepCert}

\end{document}

%% file: sections/abstract.tex
\begin{abstract}
We introduce \acronym, a tool-supported method for verifying the robustness of deep neural network (DNN) image classifiers to \emph{contextually relevant perturbations} such as blur, haze, and changes in image contrast.
While the robustness of DNN classifiers has been the subject of intense research in recent years, the solutions delivered by this research focus on verifying DNN robustness to small 
perturbations in the images being classified, with perturbation magnitude measured using established $L_p$ norms. This is useful for identifying potential adversarial attacks on DNN image classifiers, but cannot verify DNN robustness to contextually relevant image perturbations, which are typically not small when expressed with $L_p$ norms. \acronym\ addresses this underexplored verification problem by supporting:   
(1)~the encoding of real-world image perturbations; 
(2)~the systematic evaluation of contextually relevant DNN robustness, using both testing and formal verification; (3)~the generation of contextually relevant counterexamples; and, through these, (4)~the selection of DNN image classifiers suitable for the operational context (i)~envisaged when a potentially safety-critical system is designed, or (ii)~observed by a deployed system. We demonstrate the effectiveness of \acronym\ by showing how it can be used to verify the robustness of DNN image classifiers build for two benchmark datasets (`German Traffic Sign' and `CIFAR-10') to multiple contextually relevant perturbations.

\keywords{Deep neural network robustness \and Deep neural network verification \and Contextually relevant image perturbations}
\end{abstract}

%% file: sections/introduction.tex
\section{Introduction}

Deep neural network (DNN) image classifiers are increasingly being proposed for use in safety critical applications~\cite{gauerhof2020assuring,mitani2020detection,picardi2020assurance,tabernik2019deep}, where their accuracy is quoted as close to, or exceeding, that of human operators~\cite{de2018clinically}. It has been shown, however, that when the inputs to the classifier are subjected to small perturbations, even highly accurate DNNs can produce erroneous results~\cite{GoodfellowSS14,grosse2017statistical,yuan2019adversarial}. This has lead to intense research into verification techniques that check whether a DNN is robust to perturbations within a small distance from a given input, where this distance is measured using an $L_p$ norm (e.g., the Euclidean norm for $p=2$)~\cite{DuttaJST18,KaBaDiJuKo17Reluplex,katz2019marabou,PuTa10}.
These techniques are particularly useful for identifying potential adversarial attacks on DNNs~\cite{GoodfellowSS14,kurakin2016adversarial,Moosavi-Dezfooli16,PapernotMJFCS16}. They are also useful when small  changes in the DNN inputs correspond to meaningful changes in the real world, e.g., to changes in the speed and course of an aircraft for the ACAS Xu DNN verified in~\cite{KaBaDiJuKo17Reluplex}.

For DNN image classifiers, small $L_p$-norm image changes are not always meaningful. Changes that may be more meaningful for such DNNs (e.g., image blurring, hazing, variations in lighting conditions, and other natural phenomena) can also cause misclassifications, but are difficult to map to small pixel variations 
\cite{hamdi2020towards,mohapatra2020verifying}, and thus cannot be examined using traditional DNN verification techniques. What is needed for the comparison and selection of DNN image classifiers used in safety-critical systems is a \emph{contextually relevant robustness verification} method capable of assessing the robustness of DNNs to these real-world  phenomena~\cite{ashmore2019assuring,Carlini017,TianPJR18,zhang2018deeproad}. Moreover, this verification needs to be performed at DNN level (i.e., across large datasets with imagine samples from all relevant classes) rather than for a single sample image.

The tool-supported \acronym\footnote{\underline{Deep} neural network \underline{C}ont\underline{e}xtual \underline{r}obus\underline{t}ness} method introduced in our paper addresses these needs by enabling:

\vspace*{-1.75mm}
\begin{enumerate}
    \item The formal encoding of contextually relevant image perturbations at quantified perturbation levels $\epsilon\in[0,1]$.
    \item The verification of contextually relevant DNN robustness, to establish how the accuracy of a DNN degrades as the perturbation level $\epsilon$ increases.  \acronym\ can perform this verification using either test-based (fast but approximate) or formal verification (slow but providing formal guarantees). 
    \item The generation of contextually relevant counterexamples. These counterexamples provide engineers with visually meaningful information about the level of blur, haze, etc. at which DNN classifiers stop working correctly.
    \item The selection of DNNs appropriate for the operational context (i)~envisaged when a safety-critical system is designed, or (ii)~observed by the deployed system during operation.
\end{enumerate}

\vspace*{-1.75mm}
We organised the rest of the paper as follows. Section~\ref{sec:method} 
describes our \acronym\ verification method, explaining its encoding of contextual perturbations, and detailing how it can be instantiated to use test-based and formal verification.  Section~\ref{sec:implemetnation} presents the \acronym\ implementation, and Section~\ref{sec:experimental} describes the experiments we performed to evaluate it. Finally, Section~\ref{sec:related} discusses related work, and Section~\ref{sec:conclusions} provides a summary and outlines future research directions.

%% file: sections/method.tex
\vspace*{-2.2mm}
\section{\acronym\ verification method\label{sec:method}}

\vspace*{-2.4mm}
\subsection{Overview}

\begin{figure}[tb]
    \centering
    \includegraphics[width=0.96\linewidth]{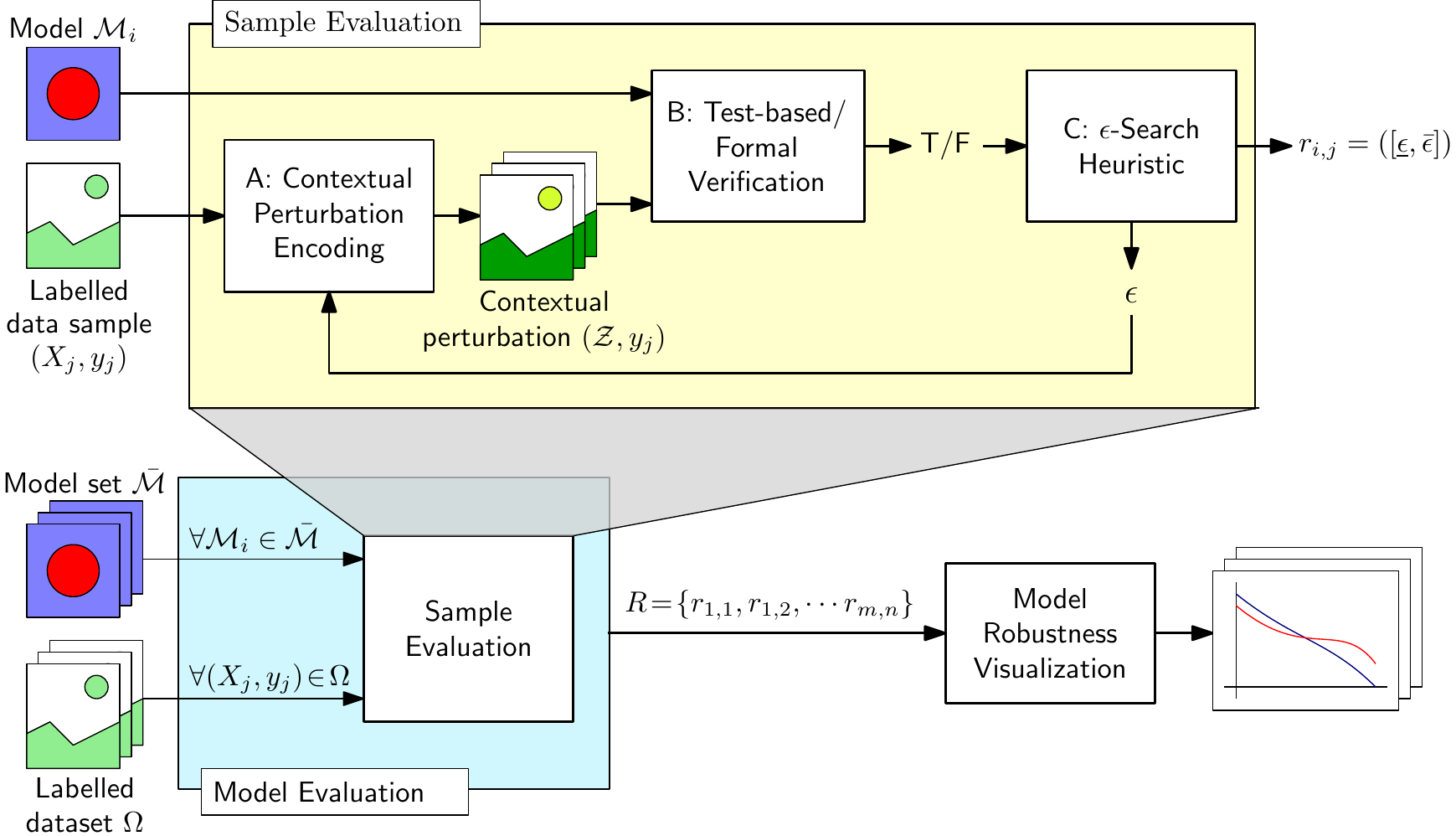}
    \caption{\acronym\ process for verifying contextually meaningful DNN robustness.}
    \label{fig:Process}
    
    \vspace*{-4mm}
\end{figure}

\vspace*{-1.5mm}
Figure~\ref{fig:Process} shows our \acronym\ method for the systematic verification of contextually relevant DNN robustness. \acronym\ accepts as input a set of $m\geq 1$ DNN models, $\bar{\mathcal{M}}$, and a dataset of $n\geq 1$ labelled image samples, $\Omega$. Each element $u \in \Omega$ is a tuple $u = (X, y)$ where $X\in \mathcal{X}$ is the input sample, $\mathcal{X}$ is the DNN input space, and $y$ is a label indicating the class into which the models should place the sample.
During model evaluation, each model $\mathcal{M}_i \in \bar{\mathcal{M}}$ is evaluated against each labelled data sample $(X_j, y_j)\in \Omega$, to find a robustness measure for that sample. The results  are then presented to the engineer as visualisations that enable model-level contextual robustness evaluation and comparison.

The sample evaluation (top of Figure~\ref{fig:Process}) is a three-stage iterative process. 
The first stage (A) encodes the contextual perturbation using a function $g:\mathcal{X}\times [0,1]\rightarrow 2^\mathcal{X}$ that maps the data sample $X_j\in\mathcal{X}$ and a \emph{perturbation level} $\epsilon\in [0,1]$ to a set of DNN inputs $\mathcal{Z} = g(X_j, \epsilon)\in 2^\mathcal{X}$ corresponding to images obtained by applying the contextual perturbation being verified (e.g., haze or blur) to the original image sample $X_j$. As we explain later in this section, $g$ applies the perturbation at level $\epsilon$ when \acronym\ employs test-based verification, and at \emph{all} levels in the range $[0,\epsilon]$ when \acronym\ employs formal verification.

The second stage (B) verifies whether the model $\mathcal{M}_i$ is robust to the contextual perturbation $(\mathcal{Z},y_j)$, i.e., whether it classifies all images from $\mathcal{Z}$ as belonging to class $y_j$. The output of this stage is a Boolean value, $\mathsf{true}$ ($\mathsf{T}$) or $\mathsf{false}$ ($\mathsf{F}$).

The final state (C) is a search heuristic that supplies the $\epsilon$ value used for the contextul perturbation encoding from stage~A, and employs binary search to identify perturbation level bounds $\underline{\epsilon},\bar{\epsilon}\in[0,1]$ such that:

\vspace*{-2mm}
\begin{itemize}
    \item either $\underline{\epsilon}<\bar{\epsilon}$, the correct class $y_j$ is predicted for $\epsilon = \underline{\epsilon}$, and a misclassification occurs for $\epsilon =\bar{\epsilon}$;
    \item or $\underline{\epsilon}=\bar{\epsilon}=0$, and the DNN misclassifies $X_j$ (with no perturbation applied).
\end{itemize} 

\vspace*{-2mm}\noindent
After checking whether $X_j$ is classified correctly by model $\mathcal{M}_i$, the search heuristic starts with $\underline{\epsilon}=0$ and $\bar{\epsilon}=1$, halves the width of the interval $[\underline{\epsilon},\bar{\epsilon}]$ in each iteration, and terminates when the width $\bar{\epsilon}-\underline{\epsilon}$ of this interval drops below a predefined value $\omega$. The final interval $r_{i,j}=[\underline{\epsilon},\bar{\epsilon}]$ is then returned.

Applying sample evaluation to each model $\mathcal{M}_i\!\in\!\mathcal{M}$ and every sample $X_j\!\in\!\Omega$ provides a result set $R = \{r_{1,1}, r_{1,2}, \cdots r_{m,n}\}$, where $r_{i,j} $ is the interval for the $i$-th model and $j$-th image sample. For each result, a counterexample $X_j'$ can be generated, if one exists (i.e., if $\underline{\epsilon}<1$), by perturbing the sample $X_j$ at level $\epsilon=\bar{\epsilon}$. Evaluating $X'_j$ using model $\mathcal{M}_i$ produces a misclassification label $\hat{y}_j$.

Visualisations of model and class robustness are then produced in which the accuracy of the models is presented as a function of the perturbation parameter $\epsilon$. By examining the accuracy of models across the range of expected perturbations, we can identify the conditions under which model switch should occur, e.g. one model may perform well at low levels of haze whilst a second may be superior as the level of haze present increases. Where the visualisations indicate that a particular class accuracy is highly sensitive to changes in $\epsilon$ this may indicate the need to choose a less sensitive model, or to gather additional training data.

\vspace*{-1.5mm}
\subsection{\acronym\ instantiation for test-based verification\label{sec:MethodTest}}

\begin{figure}[t]
    \centering
    \includegraphics[width=0.62\textwidth]{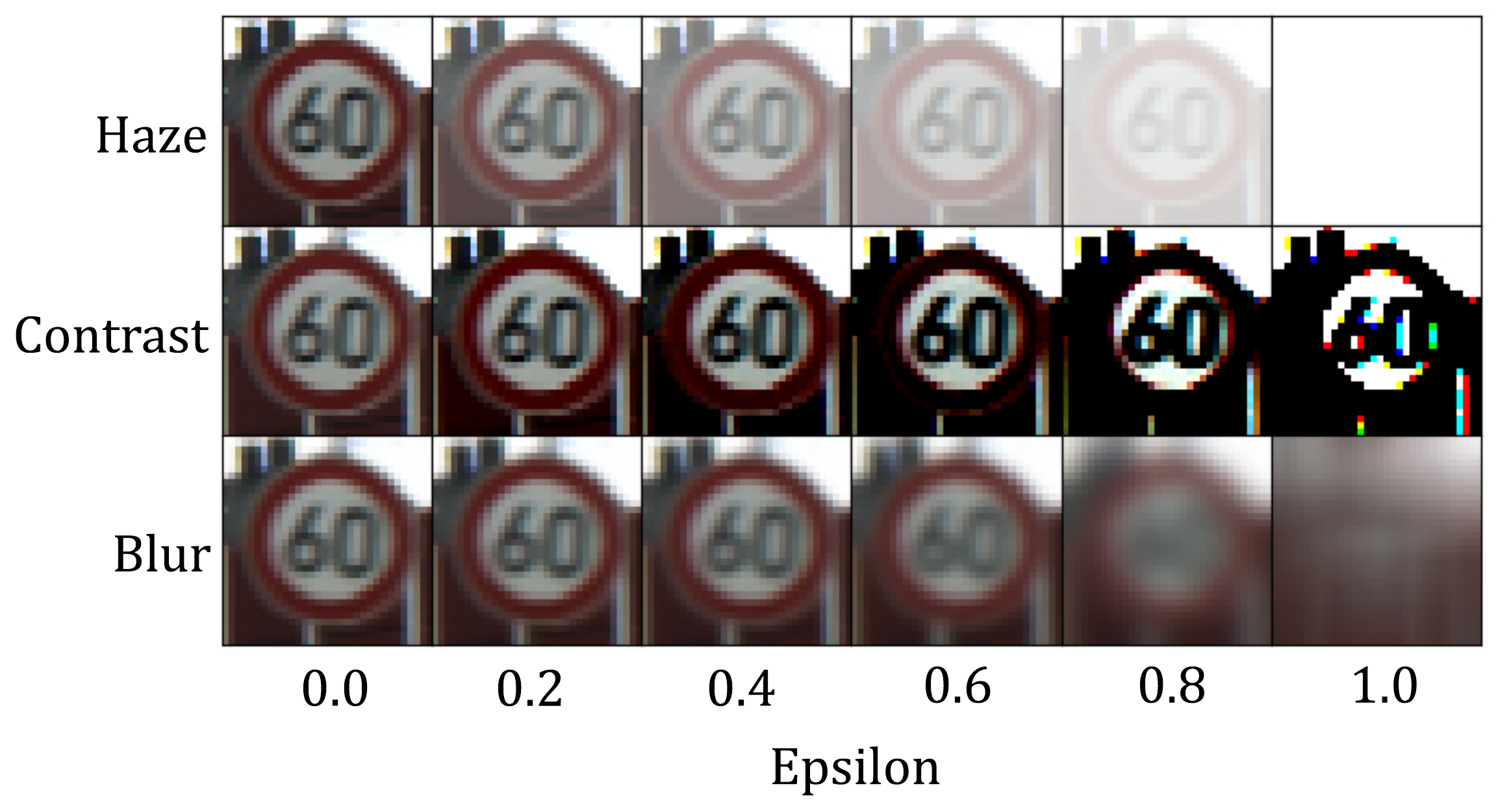}
    
    \vspace*{-2mm}
    \caption{Context perturbations applied to image sample\label{fig:epsilonImage}}
    \vspace{-5mm}
\end{figure}

\vspace*{-1.5mm}
For test-based verification,  the contextual perturbation encoding function $g$ maps an image $X$ to a set $\mathcal{Z}$ comprising a single modified image $X'$ obtained by applying a perturbation function:
\begin{equation}
  x'_{i,j} = \mathit{perturbation}(X_{i,j},\epsilon),
 \end{equation}
where $x'_{i,j}$ is the  pixel at position $(i,j$) in the modified image $X'$ and $X_{i,j}$ is  a subset of pixels from the original image $X$. 
For colour images, a sample $X$ is encoded as an array of pixels each of which is a 3-tuple of values representing the red, green and blue components of the colour in that pixel. We detail below the encoding of three typical contextual perturbations (Figure~\ref{fig:epsilonImage}).

\vspace*{1.5mm}\noindent
\textbf{Haze encoding.} 
Haze represents a phenomenon where particles in the atmosphere scatter the light reaching the observer. The effect is to drain colour from the image and create a veil of white, or coloured, mist over the image. While realistic approaches to the modelling of haze require complex models~\cite{zhang2017towards}, simplifying assumptions can be made. Assuming the haze is uniform, a haze colour may be defined as $C^{f} = (r,g,b)$ and applied to the image as:
\begin{equation}
    x'_{i,j} = (1-\epsilon) x_{i,j} + \epsilon~ C^{f}
\end{equation}
where $\epsilon \in [0,1]$ is a proxy for the density of the haze. When $\epsilon = 0$ the image is unaltered and when $\epsilon = 1$ the image is a single solid colour $C^f$. Multiplication and addition are applied to the pixel in an element-wise manner.

\vspace*{1.5mm}\noindent
\textbf{Contrast variation encoding.} 
When fixed aperture lenses are employed, or when the dynamic range of the scene is extreme, the contrast in the image may become compressed. This effect may be modelled as:
\begin{equation}
    x'_{i,j} = \textsf{Max}\left(0,\textsf{Min}\left(1,\frac{x_{i,j} - (0.5*\epsilon)}{1- \epsilon}\right) \right)
\end{equation}
The effect of applying this function is to make bright parts of the image lighter and dark parts of the image darker.

\subsubsection{Blur encoding.} 
Blurring in an image occurs when parts of the image are out of focus due to the limited capabilities of the optics employed in the system or when grease or water droplets are present on the lens. 
Blur can be synthesised using a convolutional kernel of size $2k_d + 1$ where the value of a pixel in the output image is calculated as a weighted sum of neighbouring pixels: 

\vspace*{-2.5mm}
\begin{equation}
  x'_{i,j} = \sum_{k=-k_d}^{k_d} \sum_{l=-k_d}^{k_d} \alpha_{k,l}\cdot x_{i+k,j+l}
\end{equation}

\vspace*{-1mm}\noindent
The weights $\alpha_{k,l}\in(0,1)$ are calculated by discretising a two-dimensional Gaussian curve, where the sum of weights is equal to one, $\sum_{k=-k_d}^{k_d} \sum_{l=-k_d}^{k_d} \alpha_{k,l} = 1$.
In our work, we define $\epsilon$ to be proportional to the standard deviation of the Gaussian distribution across the kernel and calculate the weights accordingly.

\vspace*{-2.5mm}
\subsection{\acronym\ instantiation for formal verification}

\vspace*{-1mm}
While test-based verification is computationally efficient, this efficiency is obtained by sacrificing completeness, i.e. if the perturbed image corresponding to an $\epsilon$ value of $p$ is not an adversarial example, we cannot guarantee that the network is robust against all perturbations with $\epsilon$ smaller than $p$. Formal verification tools, by contrast, can provide such guarantees, but typically impose constraints on the types of models and perturbations which can be analysed. 

To demonstrate the use of formal verification within \acronym, we integrated it with Marabou \cite{katz2019marabou}, a complete verification toolbox for analyzing DNNs. 
Marabou handles common piecewise linear activation functions (e.g., ReLU, Max-Pool, Sign), integrates multiple state-of-the-art bound tightening techniques \cite{eran,tjeng2017evaluating,WangPWYJ18F}, and supports parallel processing \cite{wu2020parallelization}. Given a neural network and a verification query, Marabou constructs a set of linear and piecewise linear constraints. The satisfiability of the conjunction of those constraints is evaluated using either an MILP-solver or the Reluplex procedure \cite{KaBaDiJuKo17Reluplex}. Given sufficient time, Marabou will either conclude that the query is unsatisfiable or return a satisfying assignment to the query. For this work we extended Marbou to allow for the encoding of contextual perturbations using an input perturbation function, as detailed below for haze. 

\vspace*{1.5mm}
\noindent
\textbf{Haze encoding.}  
Given a DNN model $\mathcal{M}$, an image $X$, a fog colour $C^f$, and a maximum perturbation bound $p$, we introduce variables $\vect{X}, \vect{Y}$ and $\epsilon$, denoting the DNN inputs, the DNN outputs and the perturbation bound, respectively. $\vect{X}$ has the same shape as $X$.  We then construct the following set of constraints:

\vspace*{-6mm}
\begin{subequations}
\begin{align}
\vect{Y} &= \mathcal{M}(\vect{X}) \label{eqn:haze1} \\ \displaybreak[3]
0 &\leq \epsilon \leq p \label{eqn:haze2}\\ \displaybreak[3]
\bigwedge_{i\leq |\vect{X}|} \big( \vect{x_i} &= (1-\epsilon) x_i + \epsilon~C^f \big) \label{eqn:haze3} \\ \displaybreak[3]
\bigvee_{\substack{i<=|\vect{Y}| \\ \vect{y_i} \neq \vect{y_{real}}}} \vect{y_i} & \geq \vect{y_{real}} \label{eqn:haze4}
\end{align}
\label{eqn:haze}
\end{subequations}

\vspace*{-3mm}\noindent
Checking the satisfiability of the constraints allows us to state if the network is robust against the haze perturbation for $\epsilon \leq p$. 
Constraint \eqref{eqn:haze1} denotes the relationship between $\vect{X}$ and $\vect{Y}$. It is a piecewise linear constraint if $\mathcal{M}$ only contains piecewise linear activation functions. Constraint \eqref{eqn:haze2} represents the perturbation bounds. Constraint \eqref{eqn:haze3} defines the input variables as results of the hazing perturbation. Finally, let $\vect{y_{real}}$ be the correct label, constraint \eqref{eqn:haze4} denotes that the output variable corresponding to the correct label is not greater than that of some other label. The network is locally adversarially robust against haze perturbation with $\epsilon \leq p$ if, and only if, the conjunction of the constraints above is unsatisfiable. If the constraints above is satisfiable, there exists a perturbation within $\epsilon$ such that some output other than $\vect{y_{real}}$ is maximal.

%% file: sections/implementation.tex
\vspace*{-4mm}
\section{Implementation\label{sec:implemetnation}}

\vspace*{-2mm}
We implemented our method using a Python framework which we have made available on our tool website \href{https://contextualrobustness.github.io}{https://deepcert.github.io}. The repository includes all models used in the paper, the code for the \acronym\ tool with the encoded perturbations presented in the paper, the supporting scripts required to generate the performance visualisations and instructions on how to use the framework. In addition, a version of Marabou is provided with a Python interface in which the haze perturbation from the previous section is encoded.

%% file: sections/experimental.tex
\vspace*{-2mm}
\section{Experimental Results~\label{sec:experimental}}

\vspace*{-2mm}
\subsection{Case Study 1: Road Traffic Speed Sign Classification}

\vspace*{-1mm}
Our first case study uses a subset of the German Traffic Sign benchmark~\cite{stallkamp2011german} where each sample is a 32$\times$32 RGB image. 
From this set we selected the seven classes which represented speed signs, the number of samples in each class are shown in Table~\ref{tab:cs1Data}.
We then built classification models at three levels of complexity with two models per level. The accuracy for all six models is reported in Table~\ref{tab:cs1Models} which shows accuracy increasing with model complexity. 

\begin{table}[t]
  \centering
    \caption{German Speed Sign Classification: Data and Models}
    
    \vspace*{-2mm}
    \begin{scriptsize}
    \begin{subtable}{.5\linewidth}
      \centering
        \caption{Data Sets\label{tab:cs1Data}}
    \sffamily
    \begin{tabular}{cccc}
    \toprule
         Class & Description & \# Train &\# Test  \\ \midrule
         0 & 30 km/h & 1980 & 720\\
         1 & 50 km/h & 2010 & 750\\
         2 & 60 km/h & 1260 & 450\\
         3 & 70 km/h & 1770 & 660\\
         4 & 80 km/h & 1650 & 630\\
         5 & 100 km/h & 1290 & 450\\
         6 & 120 km/h & 1260 & 450\\ \bottomrule
    \end{tabular}
    \end{subtable}%
    \begin{subtable}{.5\linewidth}
      \centering
        \caption{Models\label{tab:cs1Models}}
    \begin{tabular}{clc}
        \toprule
         Model &  Description & Accuracy\\ \midrule
         1A & \multirow{2}{*}{Small ReLu only model}  & 0.816\\ 
         1B &  & 0.847\\ \midrule
         2A & \multirow{2}{*}{Large ReLu only model}& 0.868\\ 
         2B && 0.866\\ \midrule
         3A & \multirow{2}{*}{CNN Model} & 0.988\\ 
         3B &  & 0.984\\ 
         \bottomrule
    \end{tabular}
    \end{subtable} 
    \end{scriptsize}
    
    \vspace*{-4mm}
\end{table}

\begin{figure}[b]
  \vspace*{-4mm}
  
  \centering
    \includegraphics[width=0.5\linewidth]{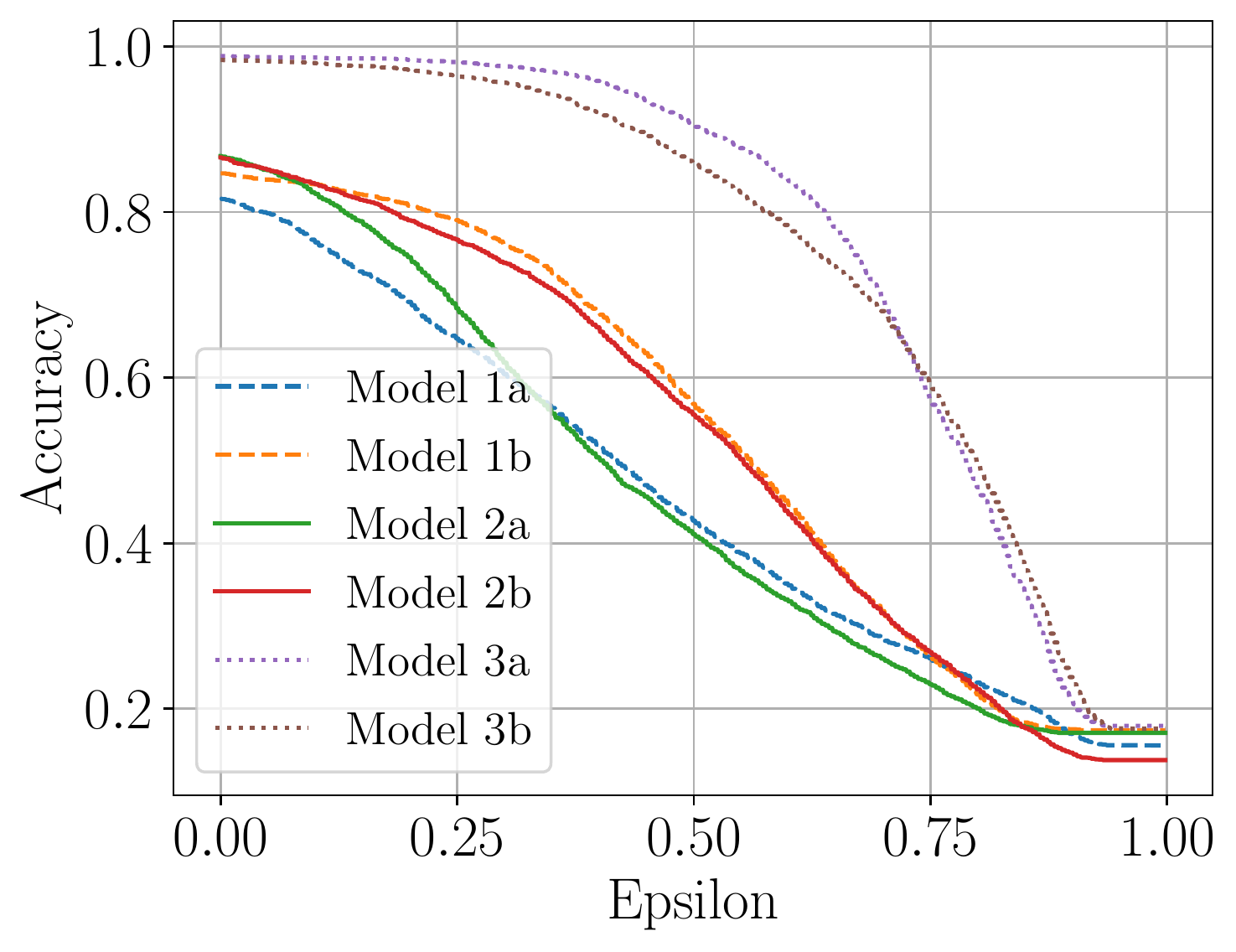}
    
  \vspace{-2mm}
  \caption{Model robustness to haze.}\label{fig:ModelHaze}
\end{figure}

\vspace*{1.5mm}\noindent
\textbf{\acronym\ with test-based verification.} For each model we applied our method using test-based verification, an initial value of $\epsilon =0.5$ and a binary search heuristic with a maximum permissible interval of 0.002. 
Figure~\ref{fig:ModelHaze} shows the impact of haze on model accuracy as $\epsilon$ is increased.
While Table~\ref{tab:cs1Models} shows model~3A to be the most accurate (0.988) without perturbation, we note that for $\epsilon \gtrapprox 0.7$, model~3B achieves superior accuracy. 
This behaviour is more clearly seen if we consider the ReLu-only models. Here model~2A has the best initial performance, but this rapidly deteriorates as $\epsilon$ increases such that other models are superior for even small amounts of haze.

These results demonstrate the dangers of selecting a model on the basis of the accuracy reported for unperturbed samples, and show how \acronym\ enables a more meaningful model selection for the operational context. Indeed, were the system to be equipped with additional sensing, to assess the level of haze present, the engineer may choose to switch between models as the level of haze increased.


\begin{figure}[t]
     \centering
     \begin{subfigure}[b]{0.49\textwidth}
         \centering
         \includegraphics[width=\textwidth]{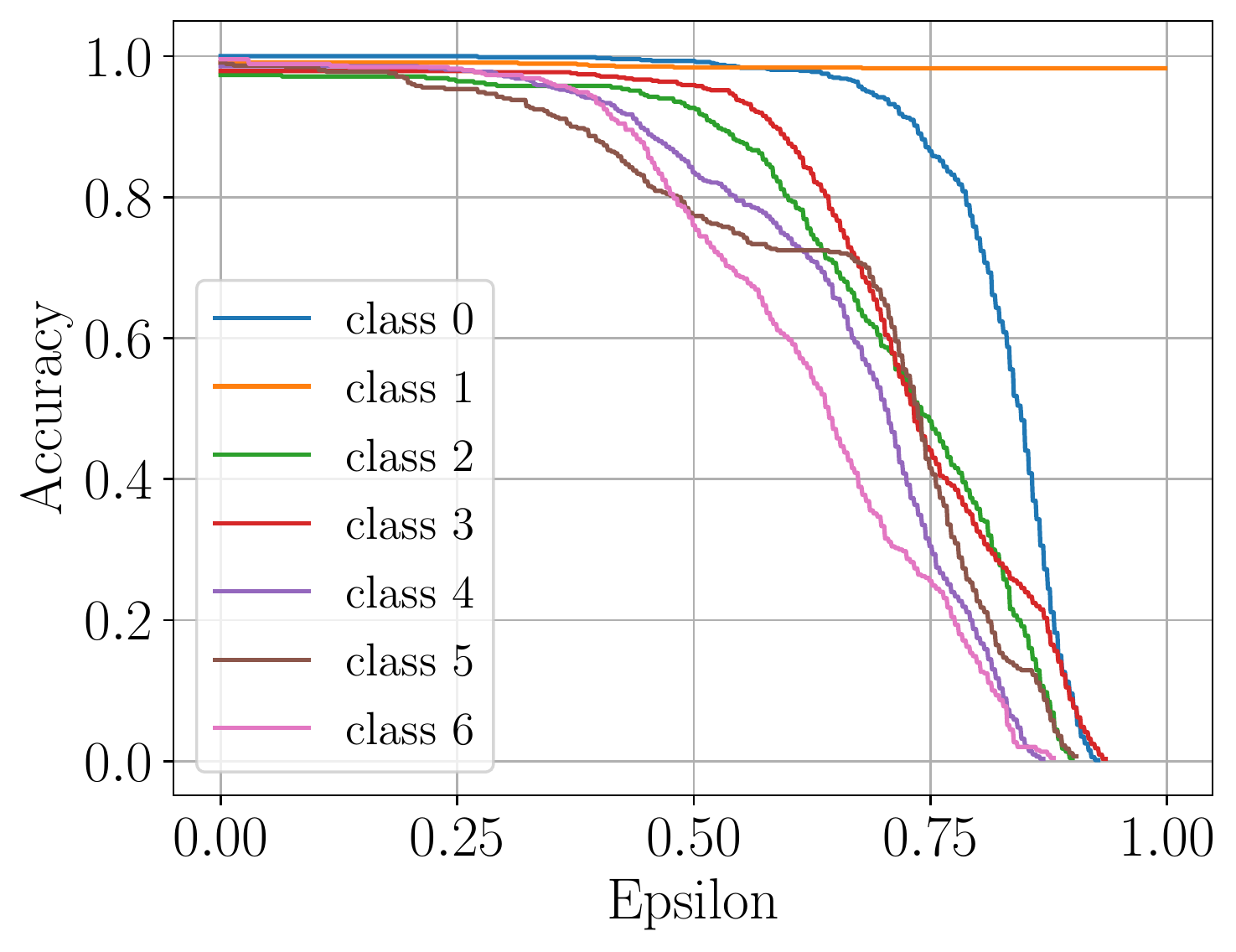}
         
         \vspace*{-1.5mm}
         \caption{Model~3A}
         \label{fig:Model5HazeClass}
     \end{subfigure}
     \hfill
     \begin{subfigure}[b]{0.49\textwidth}
         \centering
         \includegraphics[width=\textwidth]{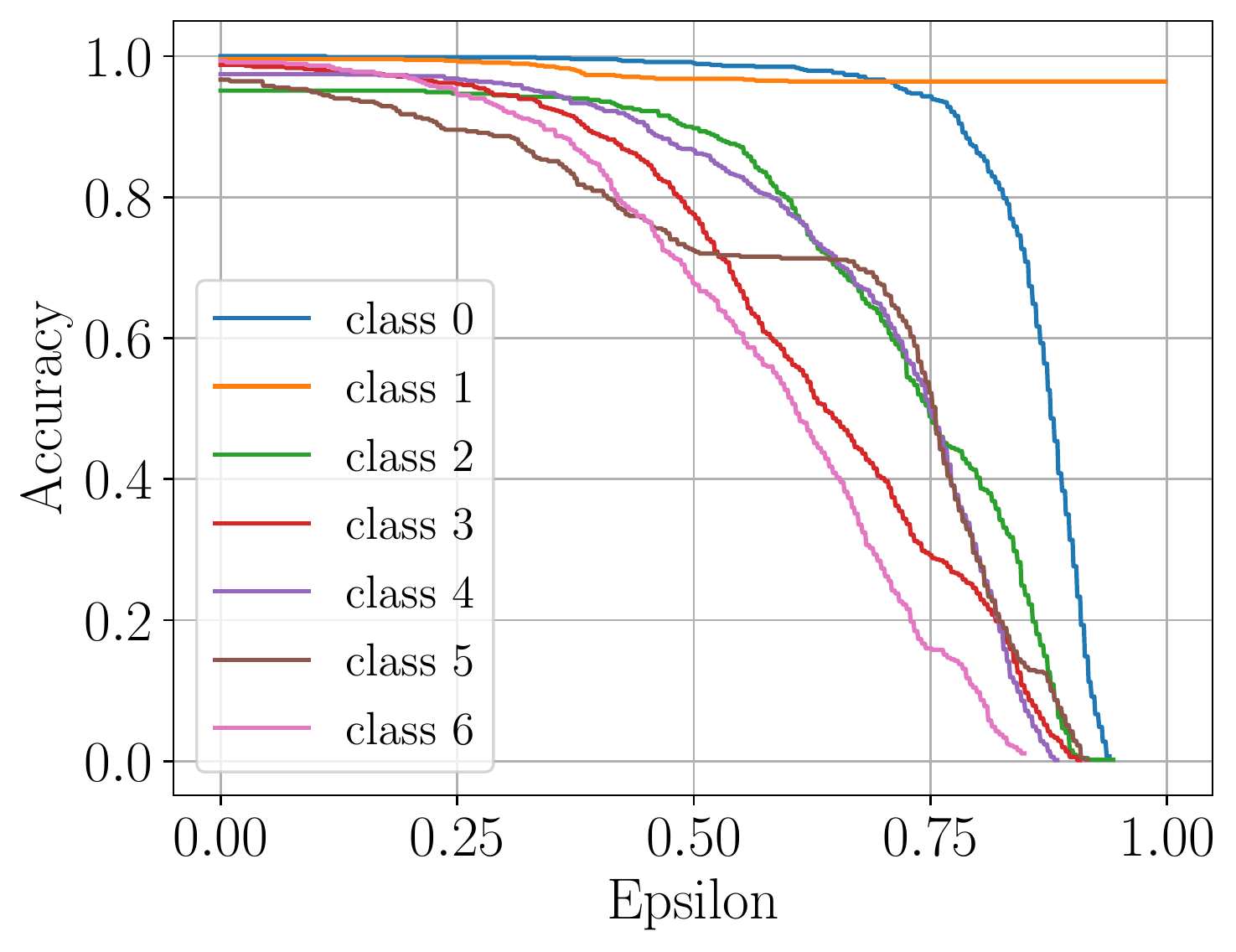}
         
         \vspace*{-1.5mm}
         \caption{Model~3B}
         \label{fig:Model6HazeClass}
     \end{subfigure}
     \begin{subfigure}[b]{0.49\textwidth}
         \centering
         \includegraphics[width=\textwidth]{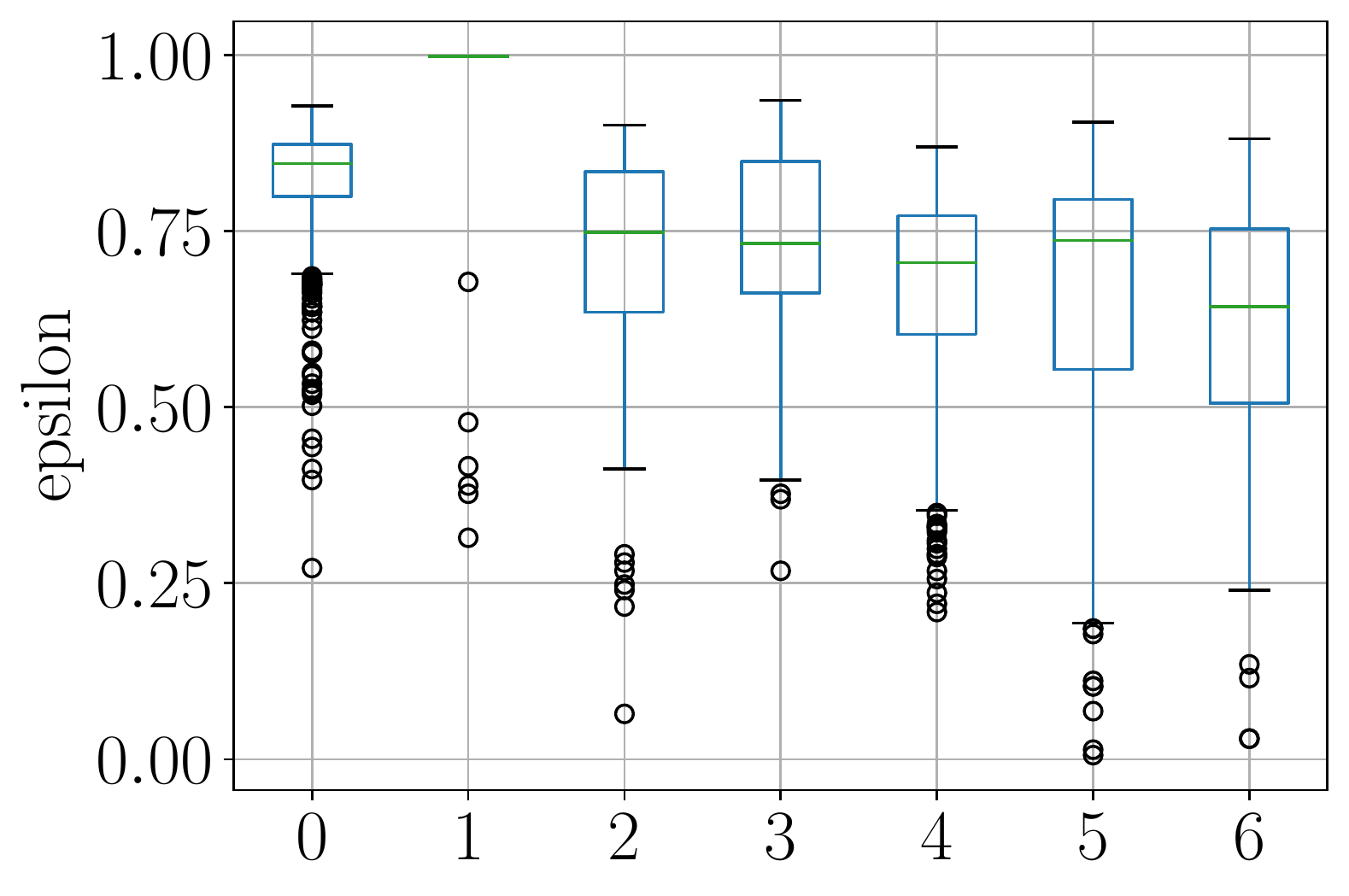}
         
         \vspace*{-1.5mm}
         \caption{Model~3A}
         \label{fig:Model5HazeClassBox}
     \end{subfigure}
     \hfill
     \begin{subfigure}[b]{0.49\textwidth}
         \centering
         \includegraphics[width=\textwidth]{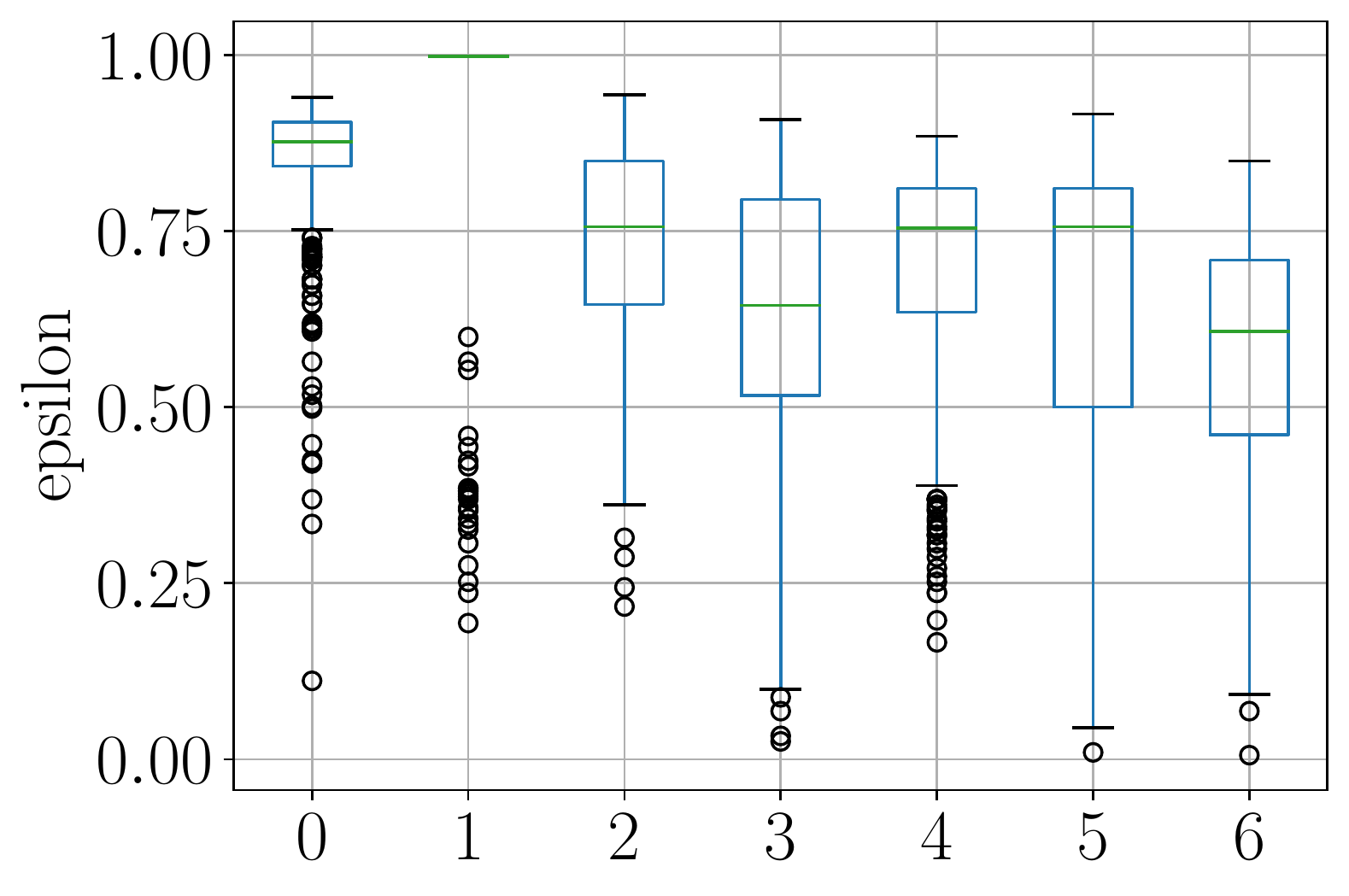}
         
         \vspace*{-1.5mm}
         \caption{Model~3B}
         \label{fig:Model6HazeClassBox}
     \end{subfigure}
     
     \vspace*{-1mm}
     \caption{Model Robustness with respect to haze\label{fig:hazeDistribution}}
     
     \vspace{-5mm}
\end{figure}

Our method also allows for the identification of those classes particularly susceptible to contextual perturbations. Figure~\ref{fig:hazeDistribution} shows the performance of the convolutional neural network (CNN) models at different levels of perturbation.  We note that class 1 is largely insensitive to haze, this is because an image perturbed with $\epsilon=1$ results in a solid colour image which is classified as class 1 by both models. For all other classes the accuracy reduces as haze increases. The amount of degradation is seen to be dependent on the sample class and the model used. For example, class 0 is more robust to haze in model 3B than in 3A with class~3 more robust in model 3A.

Figure~\ref{fig:Model5HazeClassBox} and~\ref{fig:Model6HazeClassBox} show the distribution of $\epsilon$ values required to cause misclassification. 
For class 3 we see that a number of samples are misclassified for small perturbations using model 3B but not 3A. An engineer wishing to deploy model~3B may examine these outliers to determine any correlation in image features. This may then allows for mitigation strategies at run-time or retraining with additional data samples.

\begin{figure}[t]
    \centering
    \includegraphics[width=0.95\textwidth]{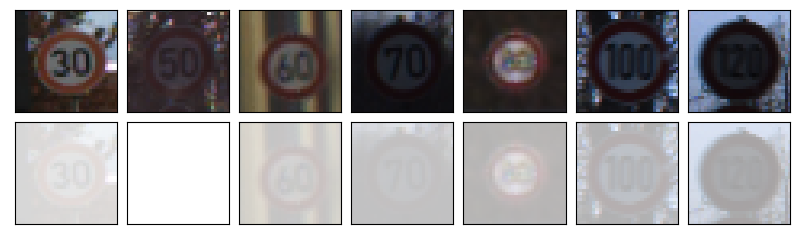}
    
    \vspace*{-2.5mm}
    \caption{Counterexamples for model~3A. Upper row is the original image, lower row has perturbation applied at the average level required for misclassification.}
    \label{fig:CE-Haze-Model5}
\end{figure}

Our method also allows for the generation of meaningful counter examples for image based classifiers. Figure~\ref{fig:CE-Haze-Model5} shows counterexamples for model~3A and illustrates the average level of haze which each class can withstand before misclassification occurs. 
This visual representation of perturbation levels allows domain experts to consider the robustness of the model with respect to normal operating conditions.

\begin{figure}[t]
     \centering
     \begin{subfigure}[b]{0.45\textwidth}
         \centering
        \includegraphics[width=\linewidth]{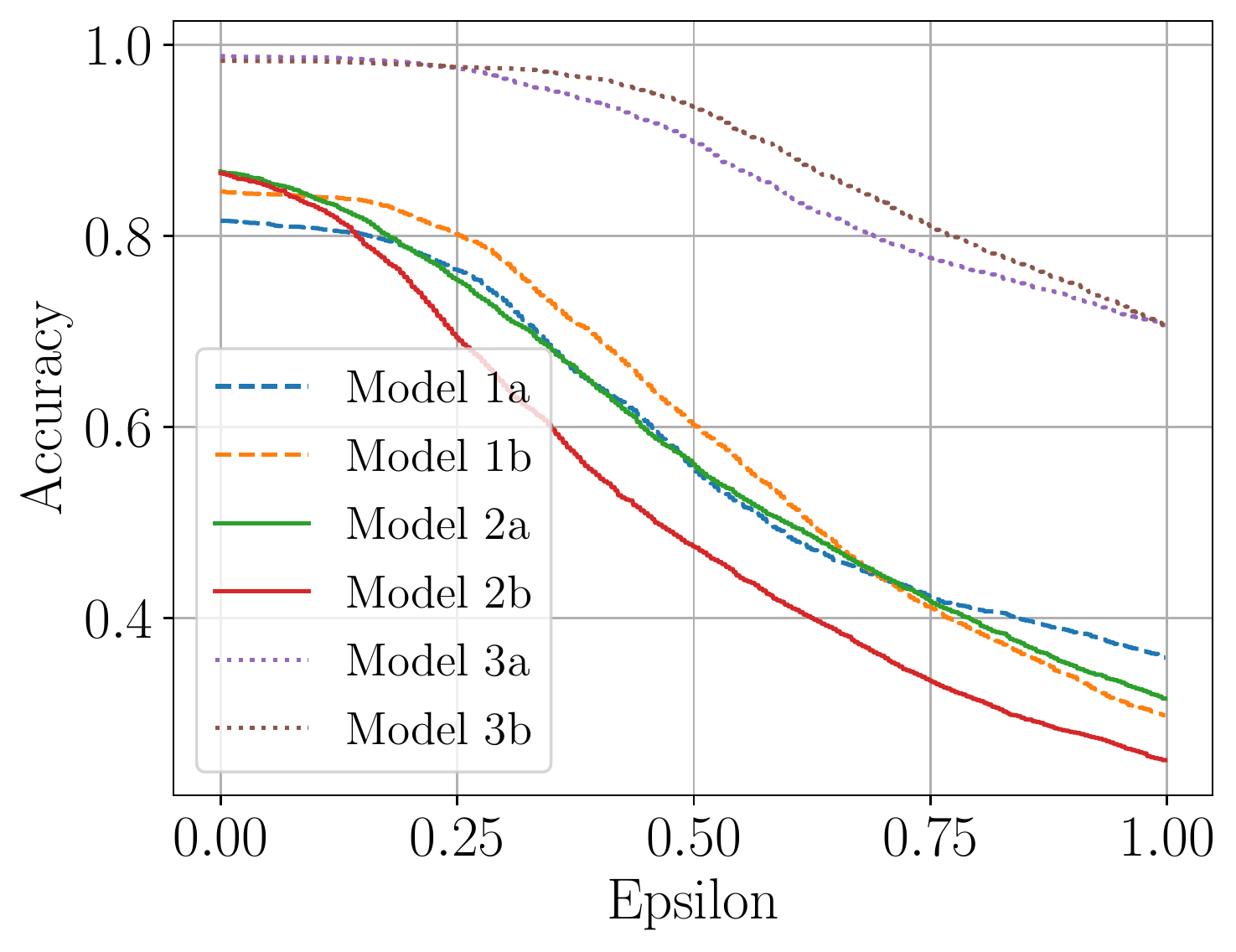}
        \caption{Contrast}
         \label{fig:ModelAccuracyContrast}
     \end{subfigure}
     \hfill
     \begin{subfigure}[b]{0.45\textwidth}
         \centering
        \includegraphics[width=\linewidth]{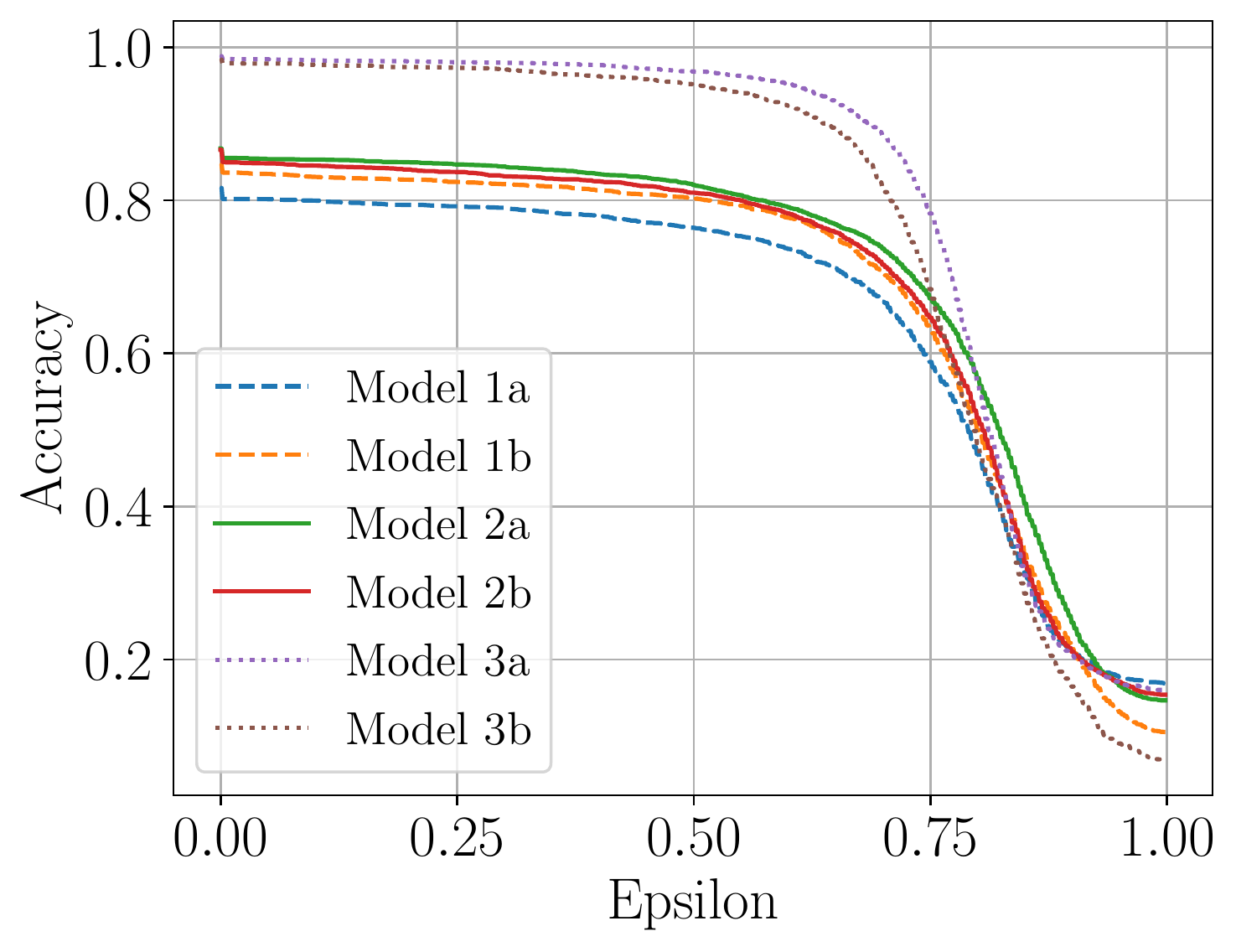}
        \caption{Blur}
         \label{fig:ModelAccuracyBlur}
     \end{subfigure}
    
    \vspace*{-1mm}
    \caption{Model accuracy with respect to increased contrast and blur effects\label{fig:ModelAccuracy2}}
     
     \vspace*{-4mm}
\end{figure}

Having demonstrated our approach using the haze perturbation we now show results for the contrast and blur effects. Model accuracy in the presence of these perturbations is shown in Figure~\ref{fig:ModelAccuracy2}. We see that whilst the accuracy of models degrades as the amount of perturbation increases, the shape of the curves and the effect on individual models is different.


Model~3A was the most accurate model for much of the perturbation range under the effects of haze, while model~3B is superior with respect to contrast effects. We also see that while model~2B was relatively robust to haze, its robustness to contrast is poor. This shows that selecting a single model for all environmental conditions is unlikely to provide optimal performance. Our method allows for a greater understanding of models weaknesses when in the presence of natural phenomena and may allow for more intelligent choices to be made.

\noindent
\textbf{\acronym\ with formal verification.} 
For model~1A and 1B, we ran our method on the first 30 images correctly classified as class 3 in the test sets to compute the minimum $\epsilon$ values for hazing using contextual perturbations and for traditional $l_{\infty}$ norm perturbations. For all 30 samples the value of $\epsilon$ found through formal verification was the same as that for the test based verification, although we can not guarantee this to be true for all samples in the testing set. 

Table~\ref{tab:verification} shows selected results from the formal verification compared with the test-based verification
Sample $\#4$ has an $l_\infty$ norm for model~1A that is lower than that of model~1B. This would indicate that model~1B is more robust. Examining  contextual robustness, however, we see that model~1A is able to withstand more haze before misclassification occurs. A similar result is shown for sample $\#52$. This time however model~1A would be judged more robust by the $l_\infty$ measure whilst model~1B is more robust according to the contextual measure.
Other samples report identical $l_\infty$ measures between models (samples 114, 47, 3 and 15) yet their response to haze is different e.g. sample $\#114$ using  model~1A is able to withstand almost twice as much haze as model~1B. 

These results demonstrate that our methods are able to use formal verification techniques, where the model form allows for such analysis. We also note that non-contextual point robustness is insufficient to assess the robustness of models in the presence of contextual perturbations.

\begin{table}[t]
\caption{Minimum $\epsilon$ values for $l_{inf}$ and hazing perturbation on test images. \label{tab:verification}}

\vspace*{1mm}
\setlength\tabcolsep{10pt}
\centering		
\sffamily
\begin{scriptsize}
\begin{tabular}{@{}ccccccc@{}}
\toprule
 & \multicolumn{3}{c}{Model 1A} & \multicolumn{3}{c}{Model 1B} \\
\cmidrule(lr){2-4} \cmidrule(lr){5-7}
& \multicolumn{2}{c}{Verification} & \multicolumn{1}{c}{Test} & \multicolumn{2}{c}{Verification} & \multicolumn{1}{c}{Test} \\
\cmidrule(lr){2-4}  \cmidrule(lr){5-7}

sample & $l_{\infty}$ & Haze & Haze  & $l_{\infty}$ & Haze & Haze  \\
\cmidrule{1-7}
4  & 0.002 & 0.623 & 0.623 & 0.006 & 0.525 & 0.525 \\
114 & 0.002 & 0.451 & 0.451 & 0.002 & 0.225 & 0.225 \\
47  & 0.006 & 0.592 & 0.592 & 0.006 & 0.752 & 0.752 \\
52  & 0.006 & 0.830 & 0.830 & 0.010 & 0.654 & 0.654 \\
3   & 0.010 & 0.764 & 0.764 & 0.010 & 0.713 & 0.713 \\
15  & 0.010 & 0.760 & 0.760 & 0.010 & 0.810 & 0.810 \\
\bottomrule
\end{tabular}

\end{scriptsize}
\end{table}

%% file: sections/cifar.tex
\subsection{Case Study 2: CIFAR-10}

\vspace*{-1mm}
In order to demonstrate that our approach is applicable to a range of problems we applied our method to a second well known classification problem, CIFAR-10. The data set consists of 60,000 32$\times$32 colour images in 10 classes with 5000 training images and 1000 test images per class. Table~\ref{tab:cifar_names} shows the names of the classes in this benchmark. The complexity and diversity of the images in this set is a more challenging classification task than the traffic sign problem. We again constructed models of increasing complexity with two models at each level. The accuracy of these models for the unperturbed test set is given in Table~\ref{tab:cifar-accuracy}.

\begin{table}[t]

\vspace*{-4mm}
    \centering
    \sffamily
    \caption{CIFAR-10 class descriptions}
    \label{tab:cifar_names}
    
    \vspace*{1mm}
    \setlength\tabcolsep{5pt}
    \begin{scriptsize}
    \begin{tabular}{ccccccccccc}
    \toprule
         class &0 &1 &2 &3 &4 &5 &6 &7 &8 &9  \\
         name &airplane & automobile & bird & cat & deer &dog &frog &horse & ship&truck \\
         \bottomrule
    \end{tabular}    
    \end{scriptsize}
    \vspace{-5mm}
\end{table}

\begin{table}[t]
    \setlength\tabcolsep{5pt}
    \centering
    \sffamily
    \caption{CIFAR-10 model accuracy \label{tab:cifar-accuracy}}
    
    \vspace*{1mm}
    \begin{scriptsize}
    \begin{tabular}{clcclcclc}
        \toprule
         \multicolumn{2}{c}{Model} & Accuracy& \multicolumn{2}{c}{Model} & Accuracy&  \multicolumn{2}{c}{Model} & Accuracy\\ \midrule
         4A & \multirow{2}{*}{Small Relu}  & 49.11&
         5A & \multirow{2}{*}{Large Relu}& 53.20&
         6A & \multirow{2}{*}{CNN} & 84.07\\
         4B &  & 47.45 &
         5B && 53.04&
         6B &  & 85.17\\ 
         \bottomrule
    \end{tabular}
    \end{scriptsize}
\end{table}

\begin{figure}[t]
     \centering
     \begin{subfigure}[b]{0.328\textwidth}
         \includegraphics[width=\textwidth]{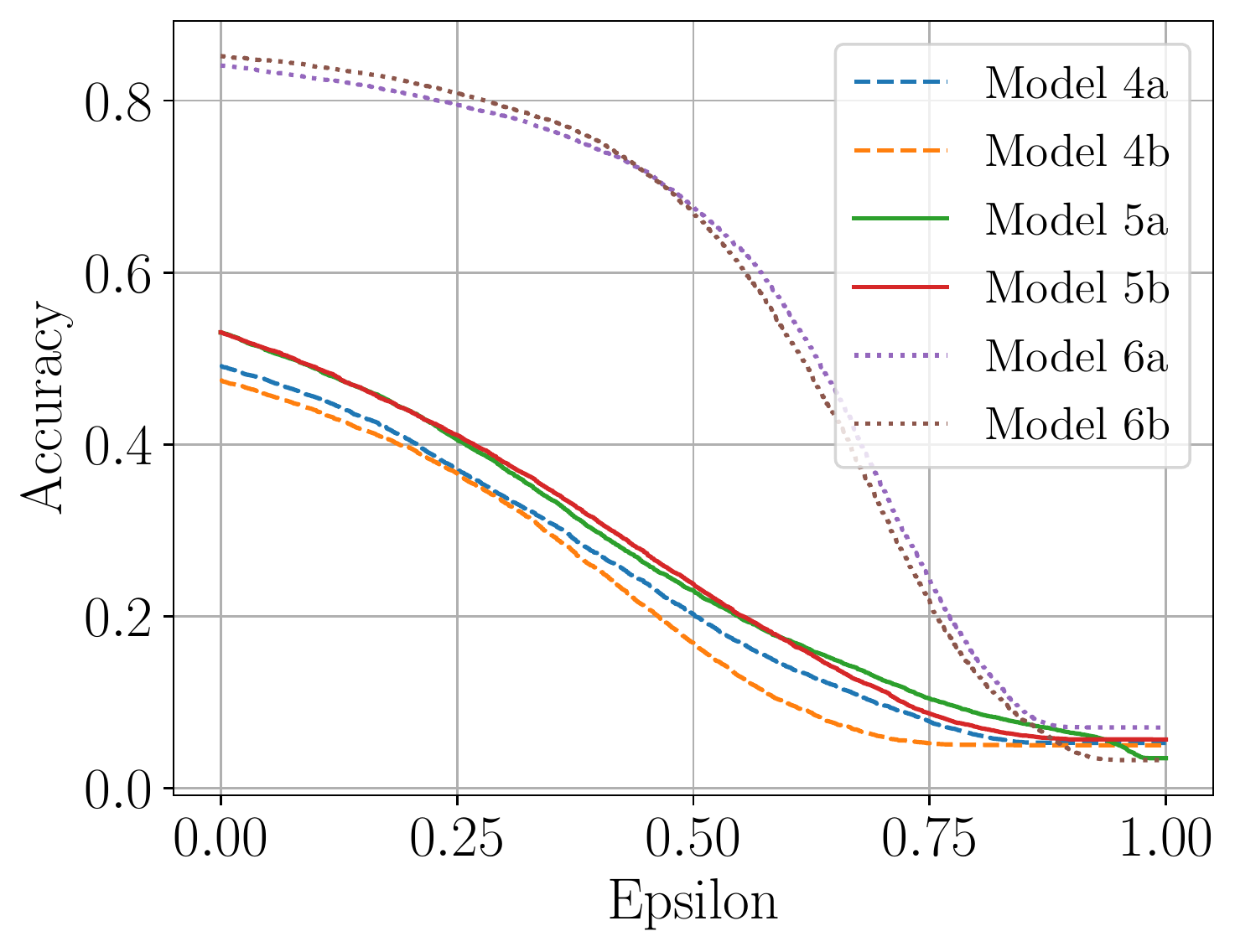}
         \caption{Haze}
         \label{fig:CifarHazeAccuracy}
     \end{subfigure}
     \begin{subfigure}[b]{0.328\textwidth}
         \includegraphics[width=\textwidth]{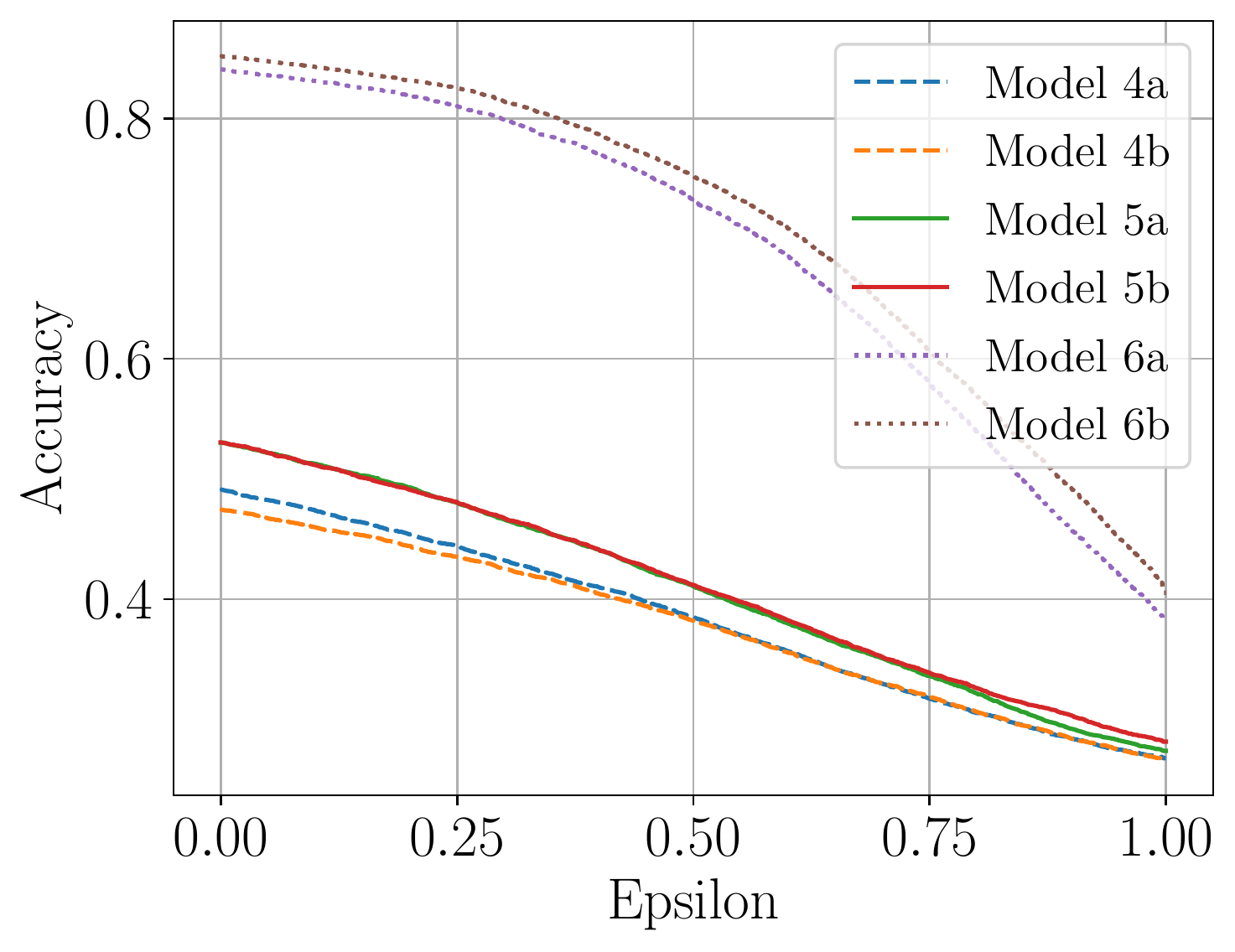}
         \caption{Contrast}
         \label{fig:CifarContrastAccuracy}
     \end{subfigure}
     \begin{subfigure}[b]{0.328\textwidth}
         \includegraphics[width=\textwidth]{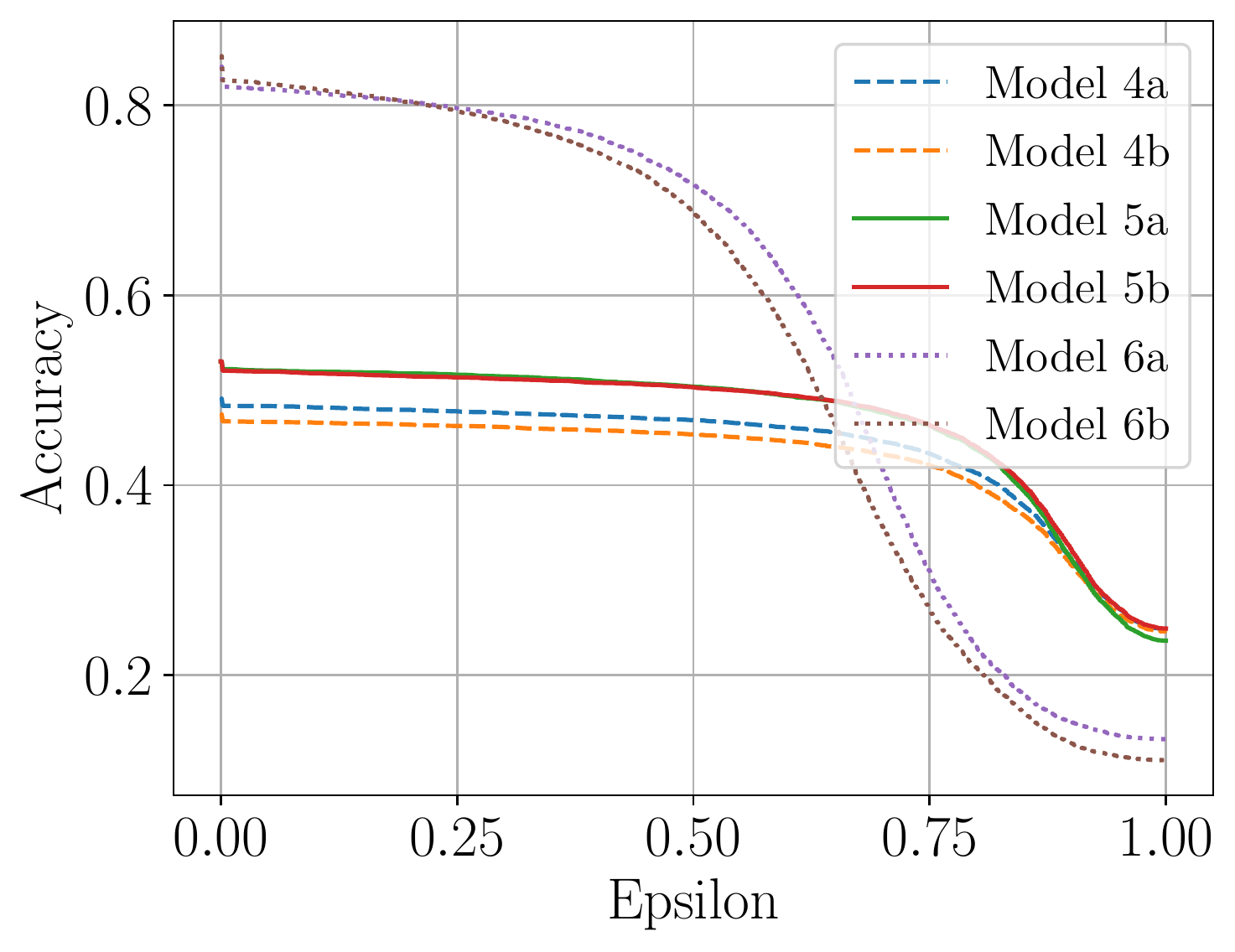}
         \caption{Blur}
         \label{fig:CifarBlueAccuracy}
     \end{subfigure}
      \caption{CIFAR-10 model robustness\label{fig:cifar-accuracy-perturbation}}
      \vspace{-5mm}
\end{figure}

\vspace*{1.5mm} \noindent \textbf{\acronym\ with test-based verification.} Model accuracy in the presence of the three forms of contextual perturbation are shown in Figure~\ref{fig:cifar-accuracy-perturbation}. We once more note the accuracy degrades as $\epsilon$ is increased for all perturbation types. For haze we observe a point at which the best model changes. This indicates that a system which is able to switch between models as the level of haze increases may demonstrate improved robustness. We also note that the CNN models outperform the simpler models by a significant margin under most conditions. For blur, however, when $\epsilon > 0.7$ the CNN models under perform the simpler models.

\begin{figure}[t]
     \centering
     \begin{subfigure}[b]{0.49\textwidth}
         \centering
         \includegraphics[width=\textwidth]{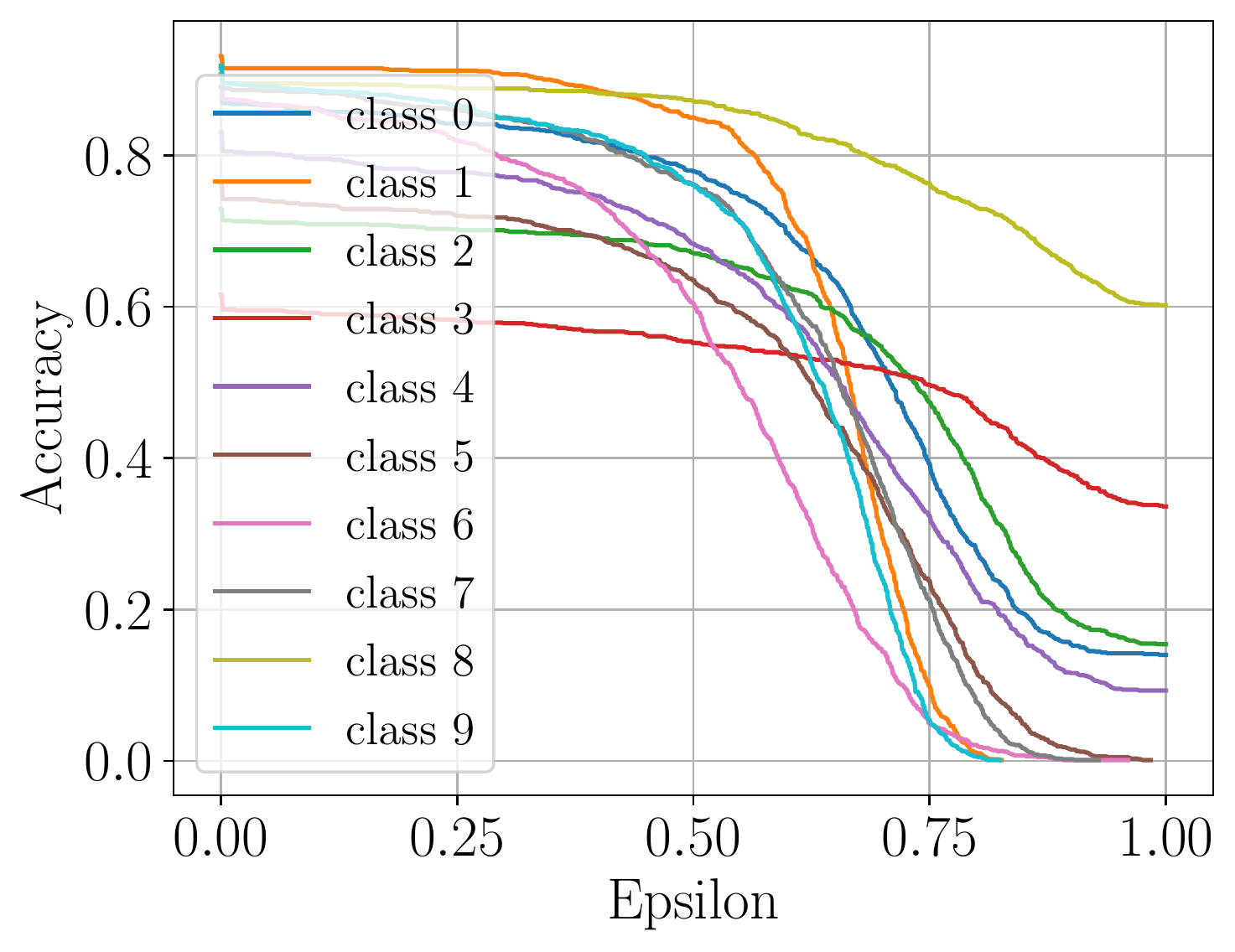}
         \caption{Model 6A}
         \label{fig:Model6ABlurClass}
     \end{subfigure}
     \hfill
     \begin{subfigure}[b]{0.49\textwidth}
         \centering
         \includegraphics[width=\textwidth]{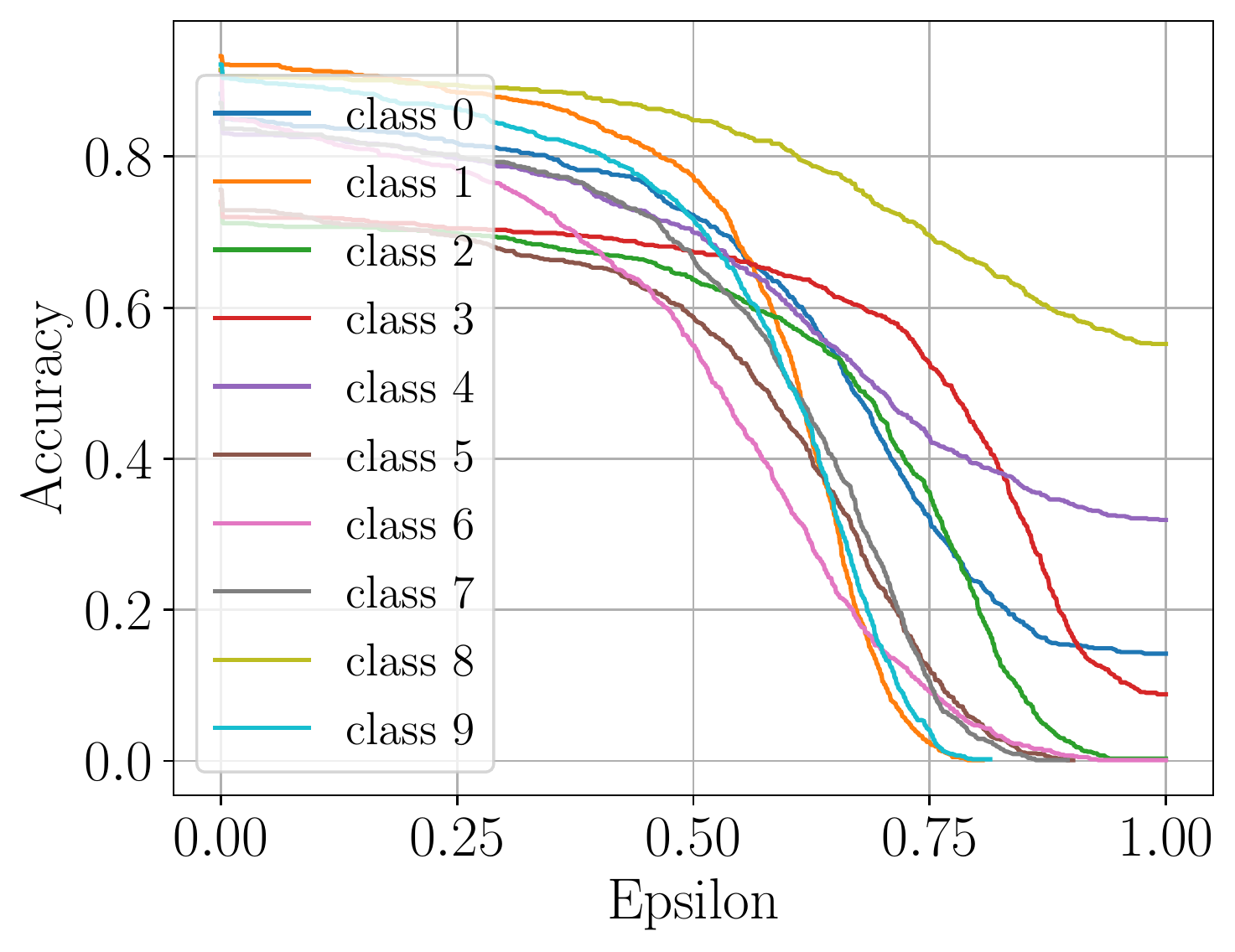}
         \caption{Model 6B}
         \label{fig:Model6BBlurClass}
     \end{subfigure}
     \begin{subfigure}[b]{0.4\textwidth}
         \centering
         \includegraphics[width=\textwidth]{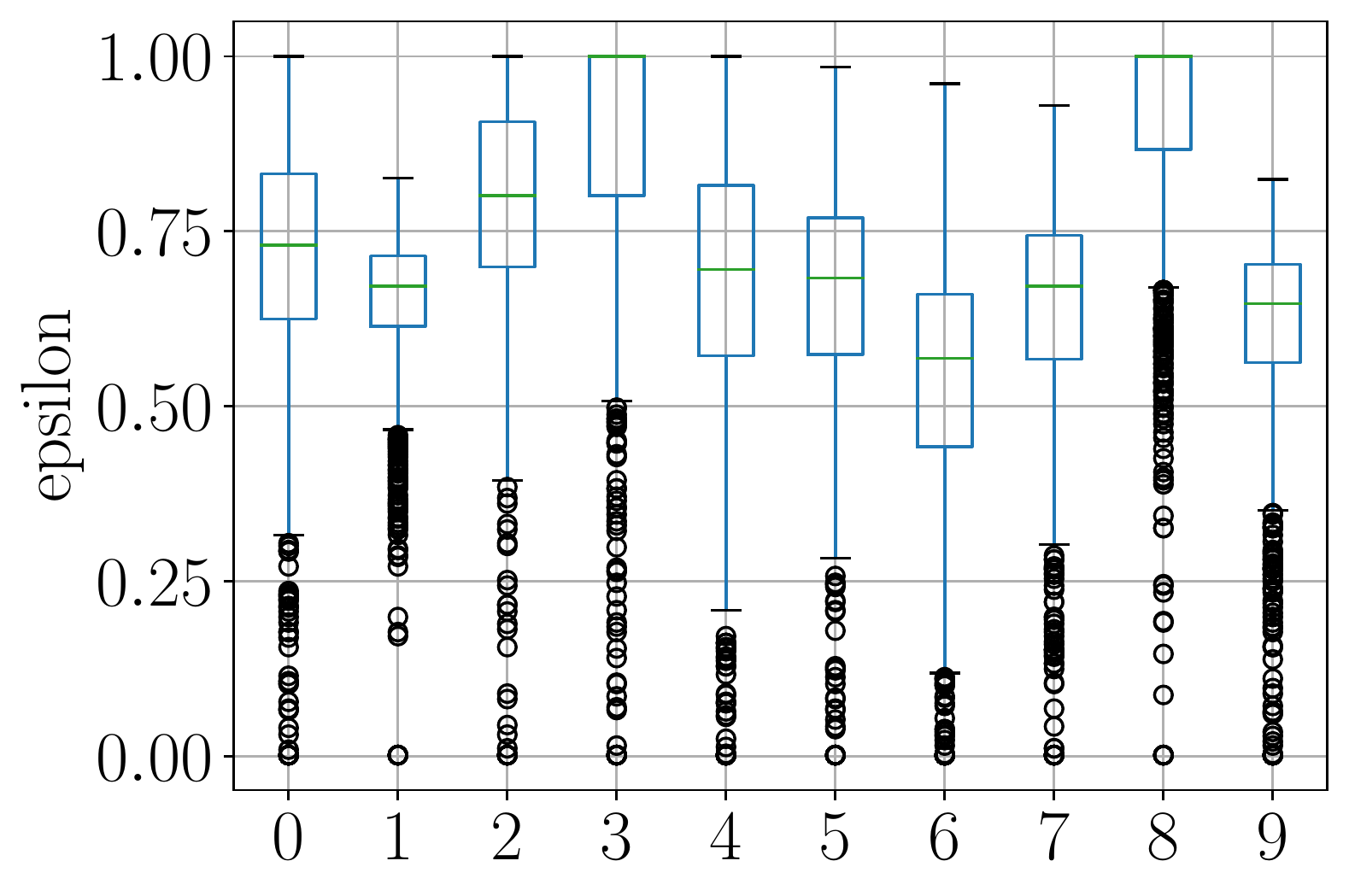}
         \caption{Model 6A}
         \label{fig:Model6ABlurClassBox}
     \end{subfigure}
     \hspace*{4mm}
     \begin{subfigure}[b]{0.4\textwidth}
         \centering
         \includegraphics[width=\textwidth]{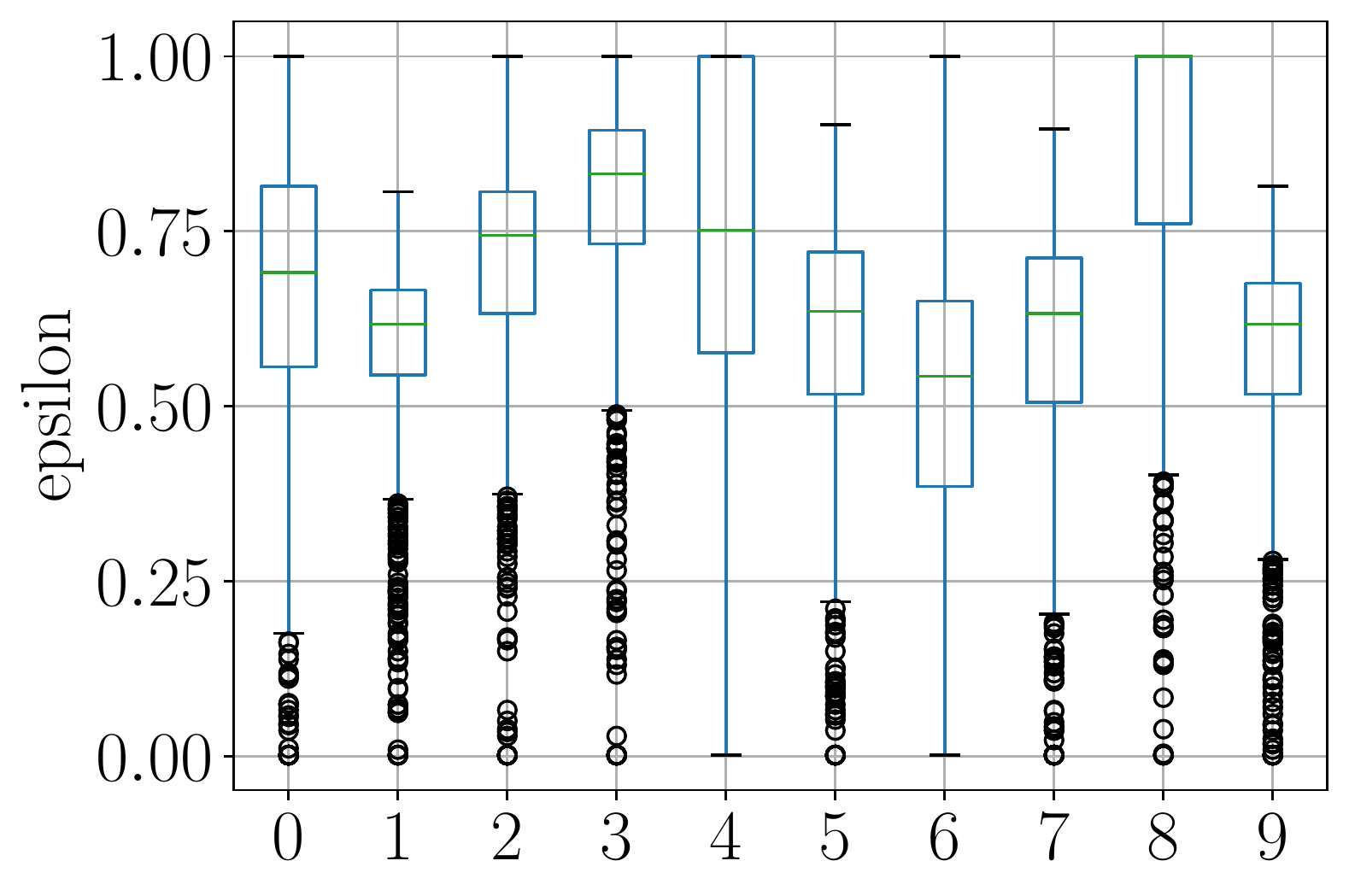}
         \caption{Model 6B}
         \label{fig:Model6BBlurClassBox}
     \end{subfigure}
     \caption{CIFAR-10 class robustness with respect to blur\label{fig:cifar-class-performance}}
     
     \vspace*{-5mm}
\end{figure}

Figure~\ref{fig:cifar-class-performance} shows the class accuracy for the CNN models subjected to the blur perturbation. We observe that the performance of classes between the models varies as shown in the traffic sign sign study.
The accuracy of class~3 in model~6A, for example, is lower than that seen in Model~6B until $\epsilon >0.7$. 

\vspace*{1.5mm}\noindent \textbf{\acronym\ with formal verification.} Formal verification was applied to models 4A and 4B by again choosing 30 samples which we perturbed with haze. The results were in line with those found for the traffic sign model, but in addition we found a sample (\#14) for model 4A which returned a lower robustness bound than when using test-based verification. Table~\ref{tab:nonLinearEpsilon} shows the predicted class $\hat{y}$ for this sample as $\epsilon$ is increased. We note that the sample is misclassified at $\epsilon=0.0723$ which was found using Marabou, it then returns to classifying the sample correctly before misclassifying again at $\epsilon=0.365$, the value found through testing. This confirms that, whilst testing may correctly identify the robustness bound for the majority of cases, formal verification is required for guarantees of robustness.

\begin{table}[t]
    \centering
    \setlength\tabcolsep{10pt}
    \sffamily
    \caption{Formal versus test-based verification, correct label $y=9$}
    \label{tab:nonLinearEpsilon}
    
    \vspace*{1mm}
    \begin{scriptsize}
    \begin{tabular}{cccc} \toprule
         $\epsilon$ & $\hat{y}$ & $\epsilon$ & $\hat{y}$  \\ \midrule
         0.002 & 9 & 0.15 & 1\\
         0.035 & 9 & 0.18 & 9\\
         0.050 & 9 & 0.2 & 9\\
         0.0723 & 1 & 0.03 & 9\\
         0.1 & 1 & 0.365 & 2 \\ \bottomrule
    \end{tabular}
    \end{scriptsize}

    \vspace{-5mm}
\end{table}

%% file: sections/related.tex
\section{Related Work\label{sec:related}}
It is well known~\cite{SzegedyZSBEGF13} that neural networks, including highly trained and smooth networks, are vulnerable to adversarial perturbations; these are small changes to an input (which are imperceptible to the human eye) that lead to mis-classifications. The vast majority of the work in this area focuses on formulating adversarial examples with respect to perturbations defined with $L_p$ norms. The problem is typically formulated as follows:  for a given network $F$ and an input $x$, find an input $x'$ for which $F(x')\neq F(x)$ while minimising $\|x-x'\|$. 

The metric used to compute the distance between points is typically the Euclidean distance ($L_2$ norm), the Manhattan distance ($L_1$ norm), or the Chebyshev distance ($L_{\infty}$ norm).  Methods for finding adversarial examples and for checking robustness of neural networks to adversarial perturbations range from heuristic and optimisation-based techniques~\cite{GoodfellowSS14,kurakin2016adversarial,PapernotMJFCS16,Carlini017,Moosavi-Dezfooli16} to formal analysis techniques which are based on constraint solving, interval analysis or abstract interpretation~\cite{HuangKWW17,KaBaDiJuKo17Reluplex,DBLP:conf/sp/GehrMDTCV18,WangPWYJ18F,WangPWYJ18E,DuttaJST18,abs-1712-06174}.
In contrast to these works, which focus on local robustness, we take a more global view, as we aim to evaluate models on many input points and use the results to assess and compare models and inform developers' choices. Furthermore, we aim to study more natural (contextual) perturbations, as we do not limit ourselves to $L_p$ norms.

Other researchers have started to look into robustness verification beyond the $L_p$-norm threat model. For instance, Semantify-NN~\cite{mohapatra2020verifying} addresses robustness verification against {\em semantic} adversarial attacks, 
such as colour shifting and lighting adjustment. It works by inserting semantic perturbation layers to the input layer of a given model, and leverages existing $L_p$-norm based verification tools to verify the model robustness against semantic perturbations. In our work, we also leverage an off-the-shelf verification tool (namely Marabou) to enable verification with respect to semantically meaningful perturbations. We do not modify the models, but instead encode the checks as Marabou queries.

%% file: sections/conclusions.tex
\section{Conclusions and Future Work\label{sec:conclusions}}

In this paper we have introduced \acronym, a tool-supported method for the systematic verification of contextually relevant robustness for neural network classifiers. We have shown that the accuracy of a DNN image classifier is a function of the perturbation type to which sample images are exposed, and that through a systematic verification of the robustness with respect to these perturbations a more informed decision may be made to select a DNN model.

In future work we plan to investigate the use of alternative formal verification techniques with \acronym, and the use of more complex models of natural phenomena, parameterised for use within the framework. We also intend to investigate methods for allowing for the systematic assessment of robustness within regions of the input space e.g. rain drops on a lens effecting part of an image.